\documentclass[conference]{IEEEtran}
\IEEEoverridecommandlockouts
\usepackage[T1]{fontenc}
\usepackage{times}

\usepackage{cite}
\usepackage{amsmath,amssymb,amsfonts}

\usepackage{algorithmic}
\usepackage{graphicx}
\usepackage{booktabs}
\usepackage{array}
\usepackage{textcomp}
\usepackage{xcolor}
\usepackage{colortbl}
\usepackage{afterpage}
\usepackage{arydshln}
\usepackage{multirow}
\usepackage{hyperref}
\usepackage{cleveref}

\def\BibTeX{{\rm B\kern-.05em{\sc i\kern-.025em b}\kern-.08em
    T\kern-.1667em\lower.7ex\hbox{E}\kern-.125emX}}
\newcommand{\mbench}{\textbf{MirrorBench}}
\newcommand{\mA}{\mathcal{A}}
\newcommand{\mD}{\mathcal{D}}
\newcommand{\CLA}[1]{{\color[HTML]{4472c4} \textbf{#1}}}
\newcommand{\CLB}[1]{{\color[HTML]{E76254} \textbf{#1}}}

\begin{document}

\title{MirrorBench: Evaluating Self-centric Intelligence in MLLMs by Introducing a Mirror}
% \thanks{Identify applicable funding agency here. If none, delete this.}
% }

% \author{\IEEEauthorblockN{
% Shengyu Guo\IEEEauthorrefmark{1}, 
% Tongrui Ye\IEEEauthorrefmark{1}, 
% Jianbo Zhang\IEEEauthorrefmark{1}, 
% Zicheng Zhang\IEEEauthorrefmark{1},
% Chunyi Li\IEEEauthorrefmark{1}
% and Guangtao Zhai\IEEEauthorrefmark{1}}

% \IEEEauthorblockA{\IEEEauthorrefmark{1}Shanghai AI Lab 
% }
% % Georgia Institute of Technology, Atlanta, Georgia 30
% % 332--0250\\
% % Email: mshell@ece.gatech.edu}
% \thanks{\textsuperscript{\dagger}Corresponding authors.}
% % \thanks{$^*$Corresponding authors.}
% }

\author{
\IEEEauthorblockN{
Shengyu Guo, 
Tongrui Ye, 
Jianbo Zhang, 
Zicheng Zhang,
Chunyi Li$^*$,
and Guangtao Zhai$^*$
}

\IEEEauthorblockA{
Shanghai AI Lab
}
\thanks{$^*$Corresponding authors.}
}

\maketitle

\afterpage{
    \begin{figure*}[t]
        \centering
        \includegraphics[width=\textwidth]{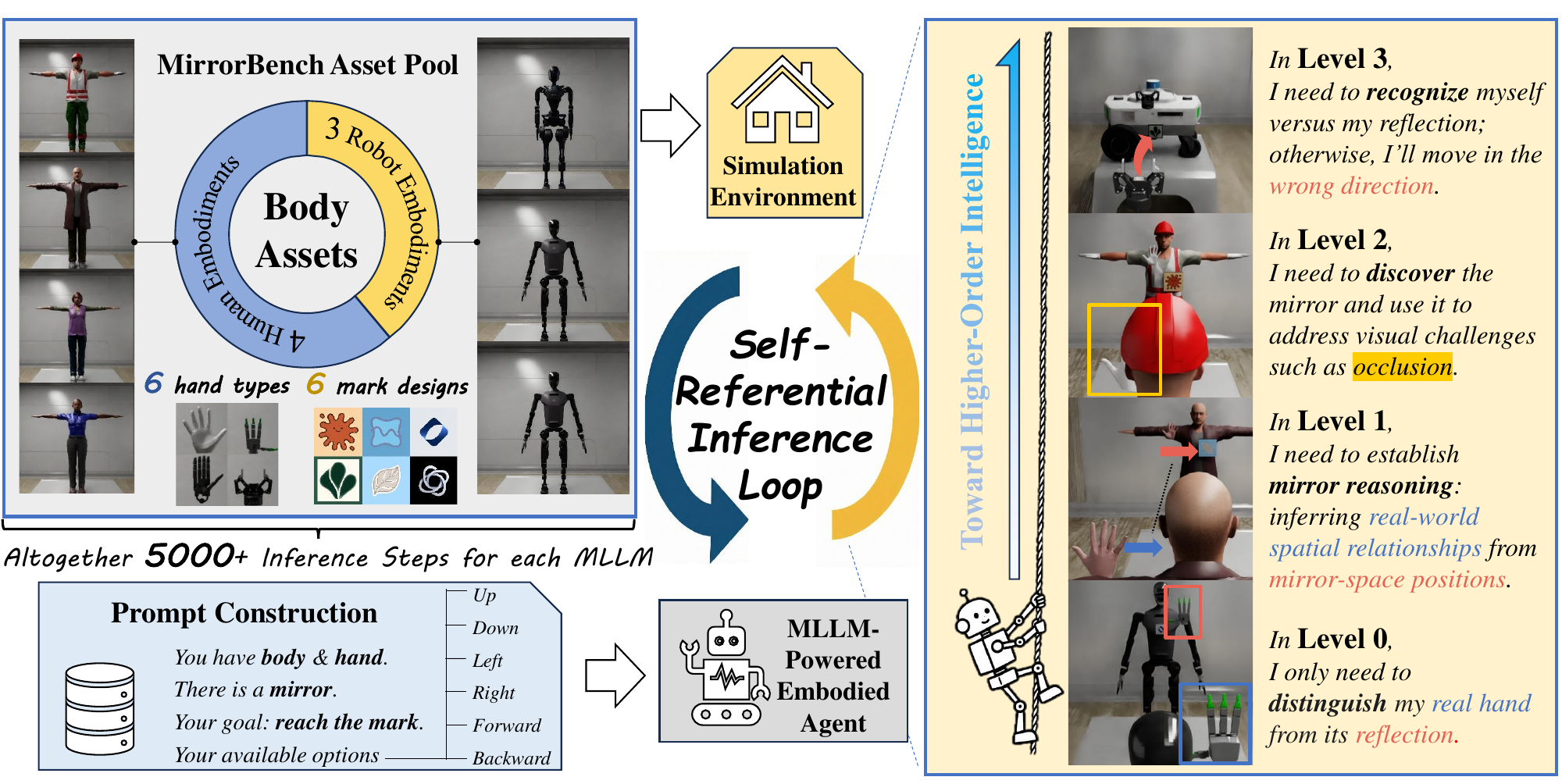}
        \vspace{-8mm}
        \caption{Overview of $\mbench$. The evaluation features a \textbf{Self-referential Inference Loop}, in which an MLLM-powered embodied agent interacts with the simulation environment. Environment is constructed from an asset pool comprising multiple body, hand and mark configurations. Prompts are generated according to the level number (bottom-left). Example scenes for Levels 0–3 (right) depict the  rising cognitive complexity toward higher-order intelligence.}
        \label{fig:fig2}
        \vspace{-6mm}
    \end{figure*}
}

\begin{abstract}
Recent progress in Multimodal Large Language Models (MLLMs) has demonstrated remarkable advances in perception and reasoning, suggesting their potential for embodied intelligence. While recent studies have evaluated embodied MLLMs in interactive settings, current benchmarks mainly target capabilities to perceive, understand, and interact with external objects, lacking a systematic evaluation of self-centric intelligence. To address this, we introduce MirrorBench, a simulation-based benchmark inspired by the classical Mirror Self-Recognition (MSR) test in psychology. MirrorBench extends this paradigm to embodied MLLMs through a tiered framework of progressively challenging tasks, assessing agents from basic visual perception to high-level self-representation. Experiments on leading MLLMs show that even at the lowest level, their performance remains substantially inferior to human performance, revealing fundamental limitations in self-referential understanding. Our study bridges psychological paradigms and embodied intelligence, offering a principled framework for evaluating the emergence of general intelligence in large models. Project page: \href{https://fflahm.github.io/mirror-bench-page/}{\texttt{mirror-bench-page}}.
\end{abstract}


\section{Introduction}
\label{sec:intro}

\begin{figure}[t!]
    \centering
    \includegraphics[width=\columnwidth]{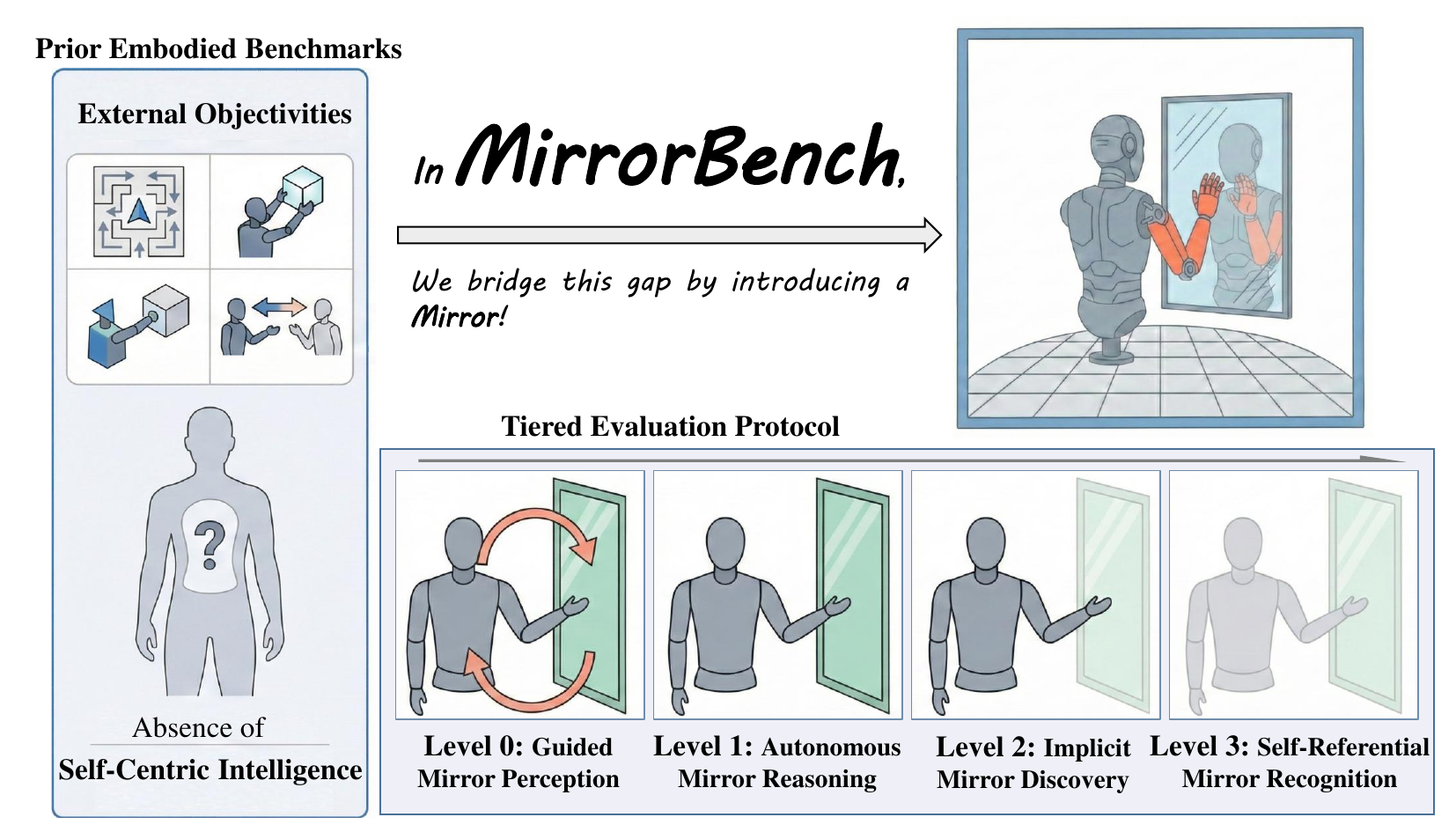}
    \vspace{-8mm}
    \caption{Existing embodied benchmarks focus on external objectives and largely neglect \textit{self-centric intelligence}. $\mbench$ bridges this gap by introducing a mirror-based setup and a \textit{tiered evaluation protocol} assessing MLLMs from basic visual perception to high-level self-representation.}
    \label{fig:fig1}
    \vspace{-6mm}
\end{figure}

Recent advances in Multimodal Large Language Models (MLLMs) have demonstrated unprecedented capabilities in visual understanding, spatial reasoning, and language grounding \cite{liu2023visual, team2023gemini, achiam2023gpt}. These developments have motivated increasing research in their application to embodied settings, where agents must perceive, reason, and act within physical or simulated environments \cite{shridhar2020alfred, li2024behavior}. To assess progress in this domain, several benchmarks have emerged, evaluating MLLMs on tasks ranging from navigation and manipulation to social interaction \cite{cheng2025embodiedeval, qiao2025navbench, zhang2025vlabench}. While these efforts provide valuable insights into the \textit{task-oriented} capabilities of embodied agents, they predominantly measure performance on \textit{external} objectives—interacting with outer space, objects or other agents—without probing the understanding of \textit{itself} within environment, as shown in \cref{fig:fig1}.

This limitation reflects a broader gap in current evaluation frameworks: the absence of \textit{self-centric intelligence} assessment of an embodied agent---namely, the understanding of its own existence, bodily boundaries, and self-to-world mappings. Despite the long-standing recognition in psychology that self-awareness and self-referential reasoning constitute fundamental aspects of intelligence \cite{duval1972theory, gallup1970chimpanzees}, embodied intelligence benchmarks largely neglect this dimension. The capacity to recognize oneself, distinguish self from others, and understand its own physical embodiment represents not merely a philosophical curiosity but a pragmatic necessity for robust, adaptive agents operating in complex, dynamic environments. Without such capabilities, models remain fundamentally limited in their ability to generalize beyond scripted interactions or handle novel scenarios requiring meta-cognitive awareness.

To bridge this gap, we introduce \mbench, a novel benchmark that adapts the classical Mirror Self-Recognition (MSR) test—a gold standard for assessing self-awareness in developmental psychology and comparative cognition \cite{gallup1970chimpanzees, amsterdam1972mirror}—to the domain of embodied MLLMs. Our framework places agents in an environment containing a mirror and a target mark on their own body, requiring them to navigate their hand to the mark using visual feedback. Crucially, $\mbench$ implements a \textit{tiered evaluation protocol} that systematically varies the amount of prior knowledge and reasoning scaffolding provided to the agent, creating four progressively challenging levels: 0) \textit{Guided Mirror Perception}, 1) \textit{Autonomous Mirror Reasoning}, 2) \textit{Implicit Mirror Discovery} and 3) \textit{Self-referential Mirror Recognition}.

This controlled progression isolates specific cognitive capabilities—from perception to self-representation—while maintaining identical physical environments across all levels, therefore enabling precise attribution of performance differences to reasoning capabilities rather than environmental complexity.

Our comprehensive evaluation of state-of-the-art MLLMs reveals striking insights. Even the strongest models perform substantially worse than human across all levels, while many open-source models—particularly smaller ones—fail to consistently outperform a random policy. These results expose a fundamental limitation in current MLLMs: the lack of robust self-referential understanding required for reasoning about embodiment in mirrored environments.

In summary, our contributions are threefold:
\begin{enumerate}
    \item We introduce \mbench, the first benchmark explicitly designed to evaluate self-centric intelligence in embodied MLLMs through a psychologically grounded mirror recognition paradigm.
    \item We propose a \textit{tiered evaluation protocol} that systematically isolates and measures specific cognitive capabilities from basic perception to high-level self-representation through controlled prompt ablation.
    \item Through extensive experiments, we reveal fundamental limitations in current MLLMs, providing crucial insights for future research in embodied intelligence.
\end{enumerate}

\afterpage{
    \begin{figure*}[t]
        \centering
        \includegraphics[width=\textwidth]{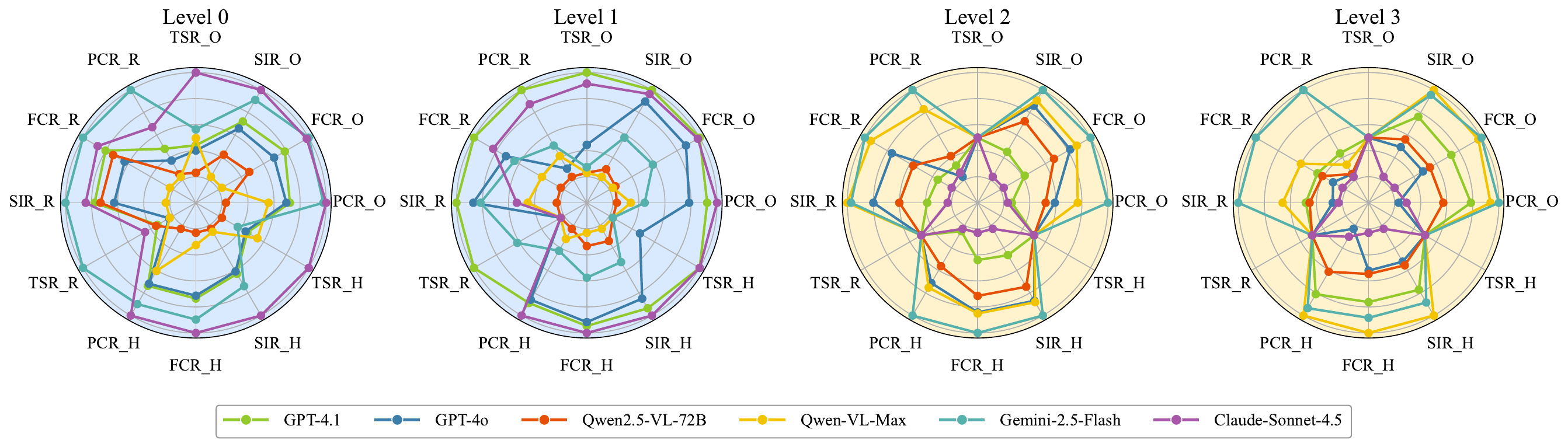}
        \vspace{-6mm}
        \caption{Radar charts showing performance of 6 representative MLLMs across 4 levels. Each  displays scores for multiple metrics under 3 settings: Overall (O), Human (H), and Robot (R). These plots illustrate how model performance varies with increasing reasoning difficulty and changing evaluation contexts.}
        \label{fig:radar}
        \vspace{-6mm}
    \end{figure*}
}

\section{Related Works}
\label{sec:related}
% \subsection{Multimodal Large Language Models}

\subsection{Evaluation for MLLMs}
Traditional evaluation benchmarks for MLLMs have predominantly focused on Visual Question Answering (VQA) tasks \cite{AIBench, zhang2025large, marino2019ok, yu2023mm, yue2024mmmu}. These benchmarks typically comprise large image–text collections, requiring models to answer questions or complete predefined tasks from a single visual observation, thereby assessing their visual understanding \cite{agrawal2019nocaps, fu2025mme} and reasoning capabilities \cite{hudson2019gqa, guo2025rbench}. However, such evaluations are inherently static and non-interactive: models generate textual responses to fixed visual inputs and bypass the perception–decision–action loop that is fundamental to agentic intelligence, preventing the assessment of decision quality, error accumulation, and behavioral consequences.

As the capabilities of MLLMs have advanced, a growing body of work has sought to incorporate interactive scenarios and tasks from embodied intelligence field into MLLM evaluation, requiring models not only to perceive and understand visual observations, but also to plan and execute actions in a goal-directed manner\cite{das2018embodied, li2024behavior, yang2025embodiedbench}. These benchmarks are typically built upon simulated environments and encompass tasks such as navigation\cite{qiao2025navbench}, object manipulation\cite{zhang2025vlabench}, and social interaction\cite{cheng2025embodiedeval}. Despite enriching the evaluation landscape of MLLMs, these efforts predominantly focus on external objectives and largely overlook the assessment of the the self-centric intelligence, thereby leaving a critical gap that $\mbench$ is designed to address.

\subsection{Mirror Self-Recognition Test}
In 1970, Gallup introduced Mirror Self-Recognition (MSR) test in chimpanzees, demonstrating their ability to recognize themselves via mirror-mediated behavior \cite{gallup1970chimpanzees}. Subsequent studies, notably by Amsterdam, showed that human infants display stage-wise progression in mirror behaviors, indicating gradual emergence of self-recognition \cite{amsterdam1972mirror}. Since then, MSR has become a foundational tool in developmental psychology and has been applied across multiple animal species.

Inspired by these findings, a line of research in robotics has explored the MSR test as an experimental paradigm for studying self-recognition in artificial agents. Existing work falls into two categories: internal reasoning approaches, examining logic- or language-driven agents \cite{govindarajulu2011towards, pipitone2021robot}, and perceptual approaches, investigating sensorimotor mechanisms supporting mirror behaviors in embodied robots \cite{steels2008robot, hoffmann2021robot}.
Across these studies, the MSR paradigm is typically simplified, or agents are endowed with strong prior knowledge, thus relaxing the conditions required for valid evaluation. In contrast, $\mbench$ systematically controls the amount of prior information provided to the agent through prompt ablation. At lower evaluation levels, agents are allowed to rely on explicit priors and guided reasoning, resembling the settings adopted in prior work. At higher levels, however, agents must succeed through their own generalization and self-referential reasoning capabilities, without access to explicit prior knowledge.

\section{MirrorBench}
\label{sec:mbench}

\subsection{Data Formulation and Construction}

A $\mbench$ scene $s$ consists of the following core components: a \textit{body}, a \textit{controllable hand}, a visual \textit{mark} attached to the body, a \textit{mirror} placed in front of the body, and associated initial poses as well as natural language descriptions.

In total, $\mbench$ features an asset pool consisting of 7 body configurations (4 human and 3 robot), 6 hand types, and 6 mark designs. This combinatorial setup yields over 5,000 inference steps for each MLLM, as shown in \cref{fig:fig2} (top-left).

\subsection{Evaluation Criteria}
Given a scene $s$, a difficulty level $\ell\in \{0, 1, 2, 3\}$, and an agent $\pi$, a single evaluation procedure is carried out as follows: 1) \textbf{Prompt Construction}: The prompt $p$ is constructed according to level $\ell$; 2) \textbf{Scene Initialization}: The simulator instantiates scene $s$ and returns the initial visual observation $o_0$ captured from a fixed camera; 3) \textbf{Agent Interaction}: At each step $i$, the agent $\pi$ receives observation $o_i$ and selects an action $a_i$ from a fixed set $\mA$ of six directional translational movements. The simulator executes $a_i$, updates the environment state, and returns the next observation $o_{i+1}$; 4) \textbf{Termination \& Evaluation}: The interaction loop terminates when either (i) the hand comes within a distance threshold of the mark (indicating \textit{task success}), or (ii) the maximum number of steps is reached (indicating \textit{task failure}). Performance is quantified using multiple metrics derived from the distance trajectory $\mD$.

The evaluation process is illustrated in \cref{fig:fig2}. The methodology for prompt construction and the design of evaluation metrics will be elaborated in \cref{subsec:levels} and \cref{subsec:metrics}.

\subsection{Tiered Evaluation Protocol}
\label{subsec:levels}
$\mbench$ introduces a four-level evaluation protocol to systematically assess the mirror-related capabilities of MLLMs. The difficulty of each level is strictly controlled through \textit{Prompt Ablation}: higher levels provide progressively less prior knowledge, thus increasing the cognitive demand on the agent while keeping the physical environment identical across all trials. This design ensures a monotonically increasing difficulty hierarchy, where success at level $\ell+1$ implies  competence at level $\ell$. The four levels are defined as follows:

\textbf{Level 0 – Guided Mirror Perception}:  
    The agent is explicitly informed that \textit{``a mirror is present in the environment''} and is provided with a full chain-of-thought (CoT) reasoning template. The task reduces to basic visual perception: distinguishing the real hand from its mirror reflection and locating the target mark within the reflected view. This level evaluates whether the MLLM can execute structured spatial reasoning when given maximal scaffolding.

\textbf{Level 1 – Autonomous Mirror Reasoning}:  
    The agent is still told that a mirror exists, but \textit{no CoT guidance is provided}. The model must autonomously infer the spatial relationship between its real hand, the mirror image, and the target object. This level tests the intrinsic ability of the model to perform mirror-aware visual reasoning without step-by-step prompting.

\textbf{Level 2 – Implicit Mirror Discovery}:   
    The presence of a mirror is \textit{not disclosed}. The agent must detect the mirror implicitly by observing visual consistencies and deduce that the observed image contains a reflection. Success requires not only perception but also meta-visual inference—recognizing that the scene includes a non-physical visual duplicate.

\textbf{Level 3 – Self-Referential Mirror Recognition:}  
    The agent receives \textit{no information} about either the mirror \textit{or} the fact that the target mark is attached to its own body. To succeed, the model must perform genuine \textit{self-referential reasoning}: it must infer that the object seen in the mirror corresponds to itself, and that moving its hand toward that visual target will result in physical contact with its own body. This level operationalizes the core challenge of the classic MSR test and represents the highest cognitive demand in our benchmark.

The progression from Level 0 to Level 3 mirrors a developmental trajectory—from scaffolded perception to autonomous self-aware action—enabling fine-grained diagnosis of MLLM capabilities in embodied mirror understanding. In the right panel of \cref{fig:fig2}, we give an intuitive demonstration of the specific requirements imposed on MLLMs at each level.

\begin{table}[t]
\centering
\caption{Definitions of metrics. Here $\mD=\{d_0,d_1,\dots,d_T\}$ denotes the Manhattan distance trajectory, and $d_{\mathrm{th}}$ denotes a threshold.}
\vspace{-2mm}
\label{tab:metrics}
\renewcommand{\arraystretch}{1}
\begin{tabular}{l c c}
\toprule
Metric (Full Name) & Abbr. & Definition \\
\midrule
Task Success Rate
& TSR
& $\displaystyle
\mathrm{TSR} =
\mathbb{I}\!\left(d_T \le d_{\mathrm{th}}\right)
$ \\

Step-wise Improvement Ratio
& SIR
& $\displaystyle
\mathrm{SIR}
= \frac{1}{T}
\sum_{i=0}^{T-1}
\mathbb{I}\!\left(d_{i+1} < d_i\right)
$ \\

Final Completion Ratio
& FCR
& $\displaystyle
\mathrm{FCR}
= 1 - \frac{d_T - d_{\mathrm{th}}}{d_0 - d_{\mathrm{th}}}
$ \\

Peak Completion Ratio
& PCR
& $\displaystyle
\mathrm{PCR}
= 1 - \frac{\min_i d_i - d_{\mathrm{th}}}{d_0 - d_{\mathrm{th}}}
$ \\
\bottomrule
\end{tabular}
\vspace{-6mm}
\end{table}

\subsection{Evaluation Metrics}
\label{subsec:metrics}

To comprehensively characterize the behavior of embodied MLLM agents in $\mbench$, we employ four complementary metrics derived from the distance trajectory between hand and mark, capturing multiple facets of task performance and providing a holistic view of both outcome and process.

We first report the \textbf{Task Success Rate (TSR)}, which measures the proportion of episodes in which the agent successfully reaches the target. TSR is the most direct  metric in embodied intelligence benchmarks, as it provides a clear indication of whether the task objective is ultimately achieved.

However, TSR is a \textit{binary, outcome-only} metric: it does not distinguish between agents that exhibit purposeful, goal-directed behavior and those that succeed by chance, nor does it provide insight into the intermediate decision-making process. To address these limitations, we introduce two additional metrics that capture complementary aspects of agent behavior.

The first is the \textbf{Step-wise Improvement Ratio (SIR)}, a process-oriented metric that decomposes the multi-step task into a sequence of single-step decisions. SIR measures the proportion of interaction steps in which the agent reduces its distance to the target. By quantifying how consistently an agent makes locally improving actions, SIR reflects the degree of sustained, directional progress toward the goal, even in episodes that do not terminate successfully.

The second category of metrics is the \textbf{Completion Ratio}, which provides a continuous measure of task progress relative to the initial distance. Specifically, the \textbf{Final Completion Ratio (FCR)} quantifies how much of the initial distance to the target has been reduced at the end of the episode. In this sense, FCR can be viewed as a continuous relaxation of TSR, capturing partial completion rather than a binary success or failure. Nevertheless, FCR remains solely dependent on the final state and does not account for transient but potentially meaningful progress achieved during the interaction.

To overcome this limitation, we further introduce the \textbf{Peak Completion Ratio (PCR)}, which measures the maximum completion achieved at any point during the episode. PCR captures the best momentary alignment with the task objective, even if such progress is not maintained until termination.

Taken together, these four metrics provide a holistic evaluation framework: TSR assesses ultimate task success, SIR characterizes step-wise decision quality, and FCR and PCR quantify global and peak task completion. This combination enables a nuanced analysis of both outcome and process. Table \ref{tab:metrics} gives the mathematic definitions of these metrics.

\begin{figure}[t]
    \centering
    \includegraphics[width=\columnwidth]{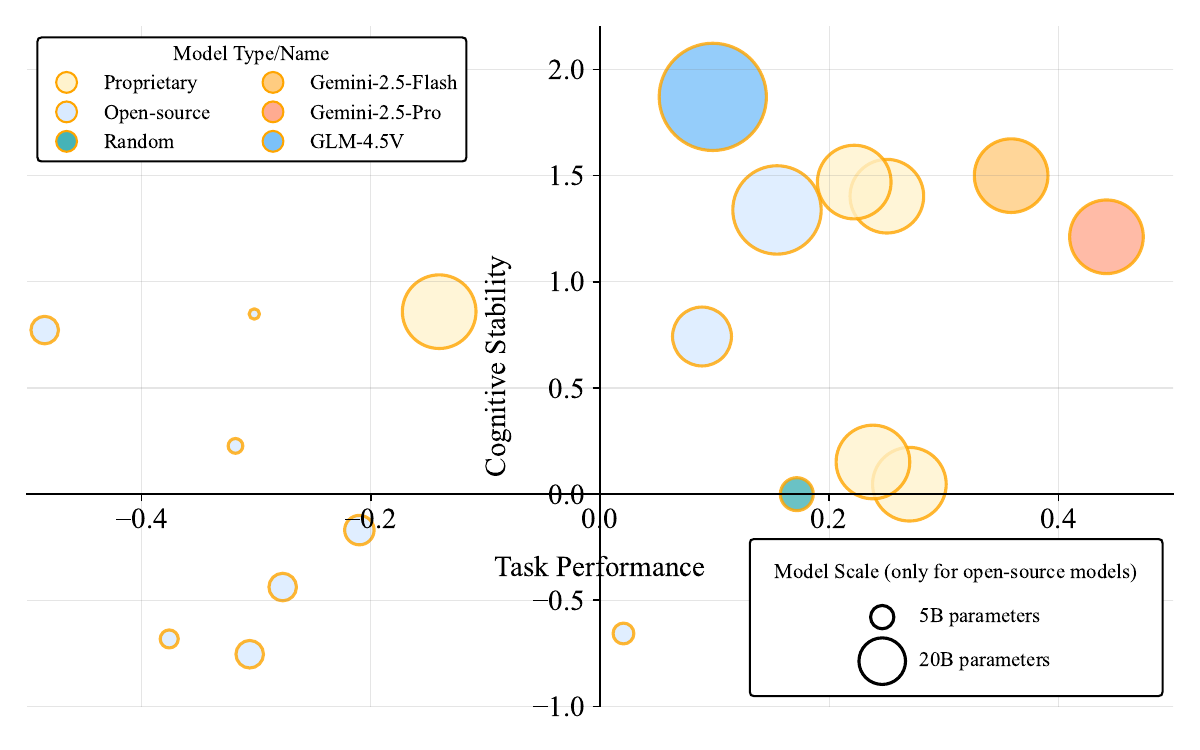}
    \vspace{-10mm}
    \caption{\textbf{Cognitive Stability} vs. \textbf{Task Performance} of MLLMs. Each point represents a model, plotted by its overall Task Performance (x-axis) and Cognitive Stability (y-axis). Cognitive Stability is computed based on the monotonicity of performance changes across levels 0–3, where higher values indicate more consistent performance degradation as task difficulty increases (see supplemental material). The bubble size encodes model scale for \textit{open-source} models, while \textit{proprietary} models are shown with a fixed size.}
    \label{fig:scatter_chart}
    \vspace{-6mm}
\end{figure}

\section{Experiments}
\label{sec:experi}

\begin{table*}[htbp]
\centering
\scriptsize
\setlength{\tabcolsep}{6pt}
\renewcommand{\arraystretch}{1.1}
\caption{$\mbench$ Results. MLLMs are ranked according to the average of the 4 metrics and 4 levels. Human agent vastly outperforms all MLLMs, while most open-source models fail to surpass the random policy. [Keys: \CLA{Highest}; \CLB{Second Highest}.]}
\label{tab:avg_level_results}
\vspace{-8pt}
\belowrulesep=0pt\aboverulesep=0pt
\begin{tabular}{
l 
c c c c c 
c c c c c 
c c c c >{\columncolor{gray!20}}c
}
\toprule
\multirow{2}{*}{\textbf{Model}}
& \multicolumn{5}{c}{\textbf{Human}} 
& \multicolumn{5}{c}{\textbf{Robot}} 
& \multicolumn{5}{c}{\textbf{Overall}} \\
\cmidrule(lr){2-6} \cmidrule(lr){7-11} \cmidrule(lr){12-16} 
& {TSR\(\uparrow\)} & {SIR\(\uparrow\)} & {FCR\(\uparrow\)} & {PCR\(\uparrow\)} & {AVG\(\uparrow\)} 
& {TSR\(\uparrow\)} & {SIR\(\uparrow\)} & {FCR\(\uparrow\)} & {PCR\(\uparrow\)} & {AVG\(\uparrow\)} 
& {TSR\(\uparrow\)} & {SIR\(\uparrow\)} & {FCR\(\uparrow\)} & {PCR\(\uparrow\)} & {AVG\(\uparrow\)} \\
\midrule
\textit{Human} & 1.000 & 0.862 & 1.000 & 1.000 & 0.966 & 1.000 & 0.895 & 1.000 & 1.000 & 0.974 & 1.000 & 0.870 & 1.000 & 1.000 & 0.968 \\ \cdashline{1-16} 
Gemini-2.5-Pro-API & \CLB{0.141} & \CLA{0.611} & \CLA{0.463} & \CLA{0.667} & \CLA{0.470} & \CLA{0.194} & \CLA{0.528} & \CLA{0.193} & \CLA{0.533} & \CLA{0.362} & \CLB{0.155} & \CLA{0.593} & \CLA{0.389} & \CLA{0.630} & \CLA{0.442} \\
Gemini-2.5-Flash-API & 0.042 & \CLB{0.584} & \CLB{0.404} & \CLB{0.607} & \CLB{0.409} & \CLB{0.125} & \CLB{0.456} & \CLB{-0.145} & \CLB{0.437} & \CLB{0.218} & 0.064 & \CLB{0.556} & \CLB{0.254} & \CLB{0.560} & \CLB{0.359} \\
GPT-4.1-API & 0.138 & 0.546 & 0.209 & 0.531 & 0.356 & 0.056 & 0.372 & -0.592 & 0.306 & 0.035 & 0.115 & 0.506 & -0.011 & 0.469 & 0.270 \\
GPT-4o-API & 0.078 & 0.551 & 0.252 & 0.504 & 0.346 & 0.015 & 0.366 & -0.659 & 0.222 & -0.014 & 0.061 & 0.509 & 0.003 & 0.428 & 0.250 \\
Qwen-VL-Max-API & 0.068 & 0.543 & 0.209 & 0.526 & 0.336 & 0.014 & 0.347 & -0.779 & 0.290 & -0.032 & 0.053 & 0.499 & -0.061 & 0.462 & 0.238 \\
Claude-Sonnet-4.5-API & \CLA{0.219} & 0.502 & 0.000 & 0.510 & 0.308 & 0.042 & 0.339 & -0.724 & 0.291 & -0.013 & \CLA{0.170} & 0.464 & -0.197 & 0.451 & 0.222 \\
\cdashline{1-16} \textit{Random} & 0.000 & 0.490 & -0.045 & 0.264 & 0.177 & 0.056 & 0.467 & -0.200 & 0.302 & 0.156 & 0.015 & 0.485 & -0.087 & 0.274 & 0.172 \\ \cdashline{1-16} 
Qwen2.5-VL-72B & 0.021 & 0.497 & -0.003 & 0.380 & 0.224 & 0.029 & 0.354 & -0.761 & 0.227 & -0.038 & 0.023 & 0.465 & -0.209 & 0.338 & 0.154 \\
GLM-4.5V-106B & 0.125 & 0.441 & -0.202 & 0.411 & 0.193 & 0.015 & 0.259 & -1.208 & 0.238 & -0.174 & 0.096 & 0.400 & -0.467 & 0.365 & 0.098 \\
Qwen2.5-VL-32B & 0.000 & 0.442 & -0.211 & 0.336 & 0.142 & 0.014 & 0.339 & -0.836 & 0.258 & -0.056 & 0.004 & 0.419 & -0.382 & 0.314 & 0.089 \\
InternVL3.5-4B & 0.005 & 0.376 & -0.436 & 0.259 & 0.051 & 0.000 & 0.329 & -0.823 & 0.229 & -0.066 & 0.004 & 0.366 & -0.538 & 0.251 & 0.020 \\
GPT-4o-Mini-API & 0.010 & 0.297 & -0.769 & 0.213 & -0.062 & 0.000 & 0.176 & -1.714 & 0.129 & -0.352 & 0.008 & 0.270 & -1.029 & 0.190 & -0.140 \\
LLaVA-1.6-Llama3-8B & 0.000 & 0.209 & -1.038 & 0.108 & -0.180 & 0.014 & 0.216 & -1.492 & 0.110 & -0.288 & 0.004 & 0.210 & -1.163 & 0.109 & -0.210 \\
LLaVA-1.6-7B & 0.000 & 0.183 & -1.320 & 0.172 & -0.241 & 0.000 & 0.176 & -1.782 & 0.115 & -0.372 & 0.000 & 0.181 & -1.445 & 0.156 & -0.277 \\
InternVL3.5-1B & 0.000 & 0.052 & -1.088 & 0.026 & -0.253 & 0.000 & 0.039 & -1.794 & 0.024 & -0.433 & 0.000 & 0.049 & -1.281 & 0.025 & -0.302 \\
Qwen2.5-VL-7B & 0.000 & 0.184 & -1.283 & 0.111 & -0.247 & 0.000 & 0.126 & -2.063 & 0.073 & -0.466 & 0.000 & 0.171 & -1.494 & 0.100 & -0.306 \\
InternVL3.5-2B & 0.000 & 0.147 & -1.239 & 0.076 & -0.254 & 0.000 & 0.100 & -2.138 & 0.073 & -0.491 & 0.000 & 0.137 & -1.484 & 0.075 & -0.318 \\
Qwen2.5-VL-3B & 0.000 & 0.068 & -1.503 & 0.052 & -0.346 & 0.000 & 0.037 & -2.027 & 0.045 & -0.486 & 0.000 & 0.063 & -1.616 & 0.050 & -0.376 \\
LLaVA-1.5-7B & 0.000 & 0.053 & -1.834 & 0.049 & -0.433 & 0.000 & 0.035 & -2.584 & 0.057 & -0.623 & 0.000 & 0.049 & -2.038 & 0.051 & -0.485 \\
\midrule
\end{tabular}
\vspace{-2mm}
\end{table*}

\begin{figure*}[h]
    \centering
    \vspace{-4pt}
    \belowrulesep=0pt\aboverulesep=0pt
    \includegraphics[width=\textwidth]{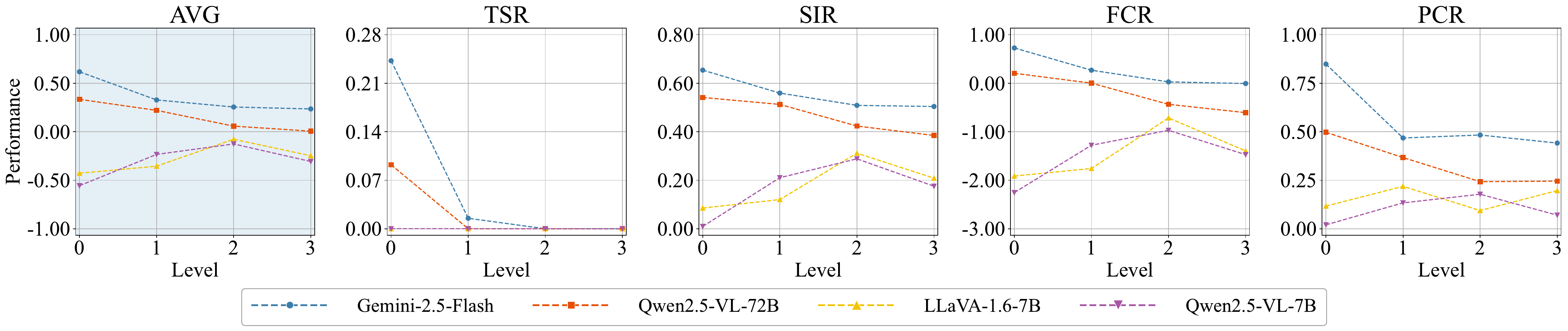}
    \vspace{-6mm}
    \caption{Metric trajectories of 4  representative MLLMs across 4 levels. For every metric, Gemini-2.5-Flash (\textit{proprietary} model) and Qwen2.5-VL-72B (large \textit{open-source} model ) exhibit a clear monotonic decline, while LLaVA-1.6-7B and Qwen2.5-VL-7B (small \textit{open-source} models) show anomalous increases. Overall, the former group consistently achieve higher scores across all levels, showing the positive relationship
between cognitive stability and task performance.}
    \label{fig:line_chart}
    \vspace{-6mm}
\end{figure*}

\begin{figure}[h]
    \centering
    
    \includegraphics[width=\columnwidth]{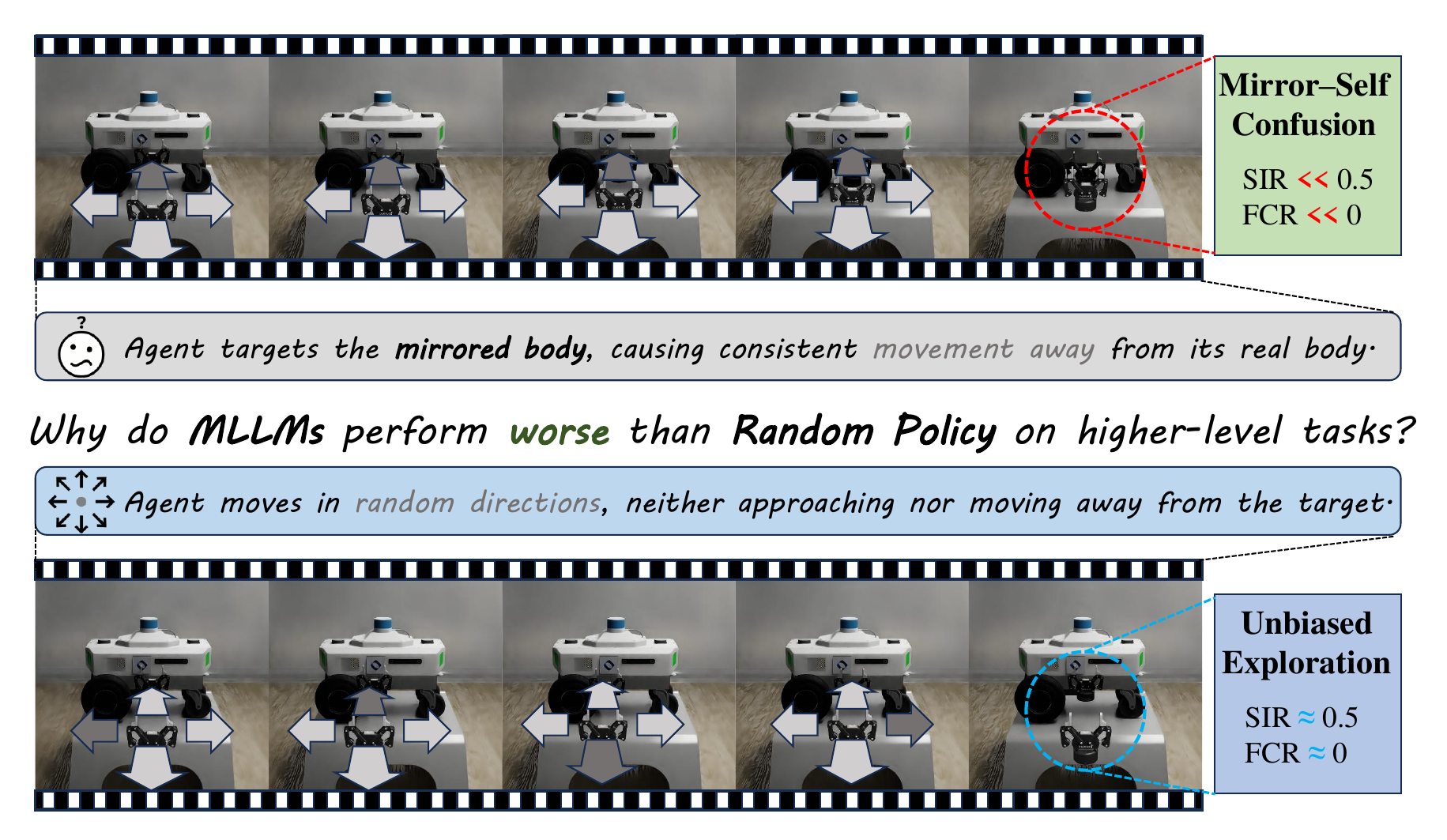}
    \vspace{-8mm}
    \caption{\textit{Mechanism of MLLMs underperformance relative to random policy.} When a MLLM falls in mirror–self confusion failure mode, it persistently targets the mirrored body rather than its physical counterpart.}
    \label{fig:fail_exp}
    \vspace{-8mm}
\end{figure}

\subsection{Benchmark Setups and Candidates}
We evaluate 18 MLLMs on \mbench, including 7 \textit{proprietary} models and 11 \textit{open-source} models. The \textit{proprietary} models include Gemini-2.5-Pro/Flash\cite{comanici2025gemini}, GPT-4o/4o-Mini\cite{openai2024gpt4o}, GPT-4.1\cite{achiam2023gpt}, Claude-Sonnet-4.5\cite{anthropic2025claudesonnet45} and Qwen-VL-Max\cite{bai2023qwenvlversatilevisionlanguagemodel}. The \textit{open-source} models include Qwen2.5-VL-3B/7B/32B/72B\cite{bai2025qwen2}, InternVL3.5-1B/2B/4B\cite{wang2025internvl3}, LLaVA-1.5-7B\cite{liu2023visual}, LLaVA-1.6-7B/Llama3-8B\cite{liu2024llavanext} and GLM-4.5V\cite{vteam2025glm45vglm41vthinkingversatilemultimodal}. Additionally, we include two reference agents: a \textit{Random} agent that uniformly samples actions, and a \textit{Human} agent that operates under the same observation and action space.

We evaluate the MLLMs under two distinct settings, \textit{Human} and \textit{Robot}, which differ in their body and hand configurations. The \textit{Human} setting adopts anthropomorphic body and hand models, whereas the \textit{Robot} setting employs robotic counterparts. All experiments are conducted in Isaac Sim, with visual observations rendered at a resolution of 1024×1024. To ensure a fairness, we uniformly set the temperature to 0 for all models. 

We set the maximum number of inference steps as the \textit{theoretical minimum steps} plus a buffer of 10 steps. For each scenario, the theoretical minimum is calculated as the Manhattan distance between the initial and target positions divided by the step size, representing the optimal path length. This adaptive threshold balances computational efficiency with the flexibility needed for the model to recover from suboptimal decisions without unnecessary resource consumption.

\subsection{Experiment Results and Discussions}
\noindent\textbf{Overall Performance: Human $>$ Proprietary MLLMs $>$ Random Policy $>$ Open-source MLLMs}
A summary of the results in \cref{tab:avg_level_results} reveals a clear \textit{performance stratification}. At the upper bound, the \textit{human} agent consistently and substantially outperforms all MLLMs, highlighting a fundamental limitation in the current capabilities of even the strongest models. Fig. \ref{fig:radar} gives a closer examination how model performance varies across levels and metrics.
The \textit{random policy} serves as a meaningful performance threshold that effectively separates \textit{proprietary} MLLMs from most \textit{open-source} models. \textit{Proprietary} models generally achieve higher scores on the majority of metrics, indicating relatively stronger self-referential reasoning and more reliable task execution, while many \textit{open-source} models—particularly those with smaller parameter sizes—struggle to consistently surpass the \textit{random baseline}. Notably, this gap persists across Human, Robot, and Overall settings, suggesting that current \textit{open-source} MLLMs remain limited in robust self-referential reasoning under mirrored embodied settings. In \cref{fig:fail_exp}, we provide in-depth explanations for why some MLLMs perform worse than \textit{random} policy.

\noindent\textbf{Across Levels: Monotonic Decline vs. Anomalous Increase}
Guided by the \textit{performance stratification} observed above, we further examine how models behave across increasing levels of difficulty. A similar \textit{two-pattern structure} emerges: \textit{Proprietary} models and large \textit{open-source} models (with higher overall performance) exhibit a clear \textit{monotonic decline} as levels increase, consistent with progressively reduced prior information and rising reasoning difficulty; In contrast, smaller \textit{open-source} models perform markedly worse and show \textit{anomalous increase}, reflecting unstable inference and weaker instruction following. This pattern is further illustrated in \cref{fig:scatter_chart}, where \textit{proprietary} models and large \textit{open-source} models cluster in the first quadrant—indicating high task performance and stable reasoning, with smaller \textit{open-source} models appearing in the opposite quadrant. The consistent monotonic trend of strong models shows a positive relationship between cognitive stability and task performance, reflecting human-like self-centric reasoning and reinforcing the benchmark’s diagnostic value. Fig. \ref{fig:line_chart} provides a more intuitive visualization of above trends.

\section{Conclusion}
\label{sec:cc}
We introduce \mbench, the first benchmark explicitly designed to evaluate self-centric intelligence in embodied MLLMs through a psychologically grounded mirror recognition paradigm. Our tiered evaluation protocol systematically isolates cognitive capabilities from perception to self-representation, revealing fundamental limitations that current MLLMs lack the  self-referential understanding necessary for adaptive behavior in complex environments. By highlighting this, we aim to inspire future research directions that address the critical gap in self-centric intelligence, guiding the development of more robust and trustworthy embodied systems.

\section*{Acknowledgment}

This work was supported by New Generation Artificial Intelligence-National Science and Technology Major Project (2025ZD0124104) in collaboration with Shanghai Artificial Intelligence Laboratory, and National Natural Science Foundation of China under Grants 625B2118, 62225112.

\bibliographystyle{IEEEtran}
\bibliography{icme2026references}

\appendices
\section{Additional Analysis}

\label{app:additional}

\subsection{Cognitive Stability}
\label{app:cmss}

Here we demonstrate the computation of Cognitive Stability score used in fig. 4, which measures the monotonicity of performance changes across levels. Let $s_{m,\ell}^{(k)}$ denote the performance score of model $m$ 
on scenario $k$ at difficulty level $\ell \in \{0,1,2,3\}$.
Cognitive Stability of model $m$ is defined as
\begin{equation}
\label{eq:cognitive_stability}
\mathrm{CS}(m)
=
\frac{1}{|\mathcal{S}|}
\sum_{k \in \mathcal{S}}
\sum_{\ell=0}^{2}
\operatorname{sign}
\!\left(
s_{m,\ell}^{(k)} - s_{m,\ell+1}^{(k)}
\right),
\end{equation}
where $\mathcal{S}$ denotes the set of evaluated scenes, and
\begin{equation}
\operatorname{sign}(x)=
\begin{cases}
\;\;\,1, & x > 0, \\
\;\;\,0, & x = 0, \\
-1, & x < 0 .
\end{cases}
\end{equation}

It should be noted that the computation of Cognitive Stability relies on a key
property of $\mbench$: the evaluation levels differ only in the amount of prior
knowledge provided in the prompt, while the physical scenario remains fixed.
It is precisely this property that allows level-wise performance variations to be
interpreted as changes in cognitive demand.

\begin{figure}[h]
    \centering
    
    \includegraphics[width=\columnwidth]{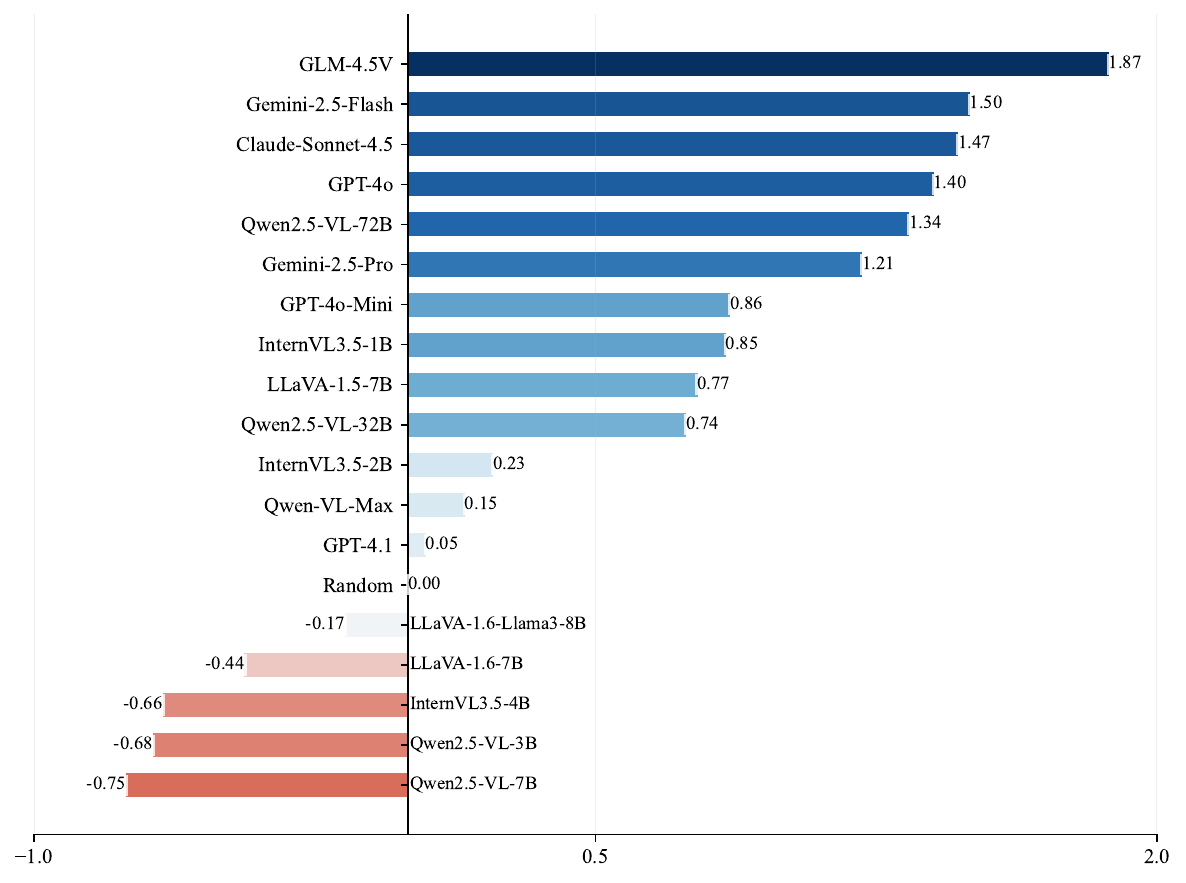}
    \vspace{-8mm}
    \caption{Cognitive Stability Ranking for MLLMs and \textit{Random} Policy.}
    \label{fig:cmss_rank}
\end{figure}

Fig. \ref{fig:cmss_rank} showcases a comparative analysis of the Cognitive Stability score between MLLMs and \textit{random} policy. Results are consistent with our previous conclusion that \textit{proprietary} models and large \textit{open-source} models exhibit higher stability.

\subsection{Human vs. Robot: Out-of-Distribution Challenges}
Based on the comprehensive evaluation across all difficulty levels, a consistent performance gap emerges between the Human and Robot settings. As shown in \cref{tab:level_setting_avg}, the average scores for the Robot setting are consistently lower than those for the Human setting across all four levels. This systematic performance degradation in robot-centric scenarios indicates significant out-of-distribution (OOD) challenges faced by current MLLMs. The results suggest that MLLMs, predominantly trained on human-centric visual and linguistic data from internet sources, exhibit limited generalization capabilities when applied to robot-involving contexts.
\begin{table}[htbp]
\centering
\small
\setlength{\tabcolsep}{10pt}
\renewcommand{\arraystretch}{1.3}
\caption{Average performance across all models by level and setting.}
\label{tab:level_setting_avg}
\begin{tabular}{lccc}
\toprule
\textbf{Level} & \textbf{Human} & \textbf{Robot} & \textbf{Overall} \\
\midrule
0 & 0.131 & -0.076 & 0.076 \\
1 & 0.128 & -0.079 & 0.073 \\
2 & -0.005 & -0.205 & -0.058 \\
3 & -0.072 & -0.371 & -0.152 \\
\midrule
\textbf{AVG} & \textbf{0.046} & \textbf{-0.183} & \textbf{-0.015} \\
\bottomrule
\end{tabular}
\vspace{2mm}
\end{table}

\subsection{Further Analysis on Level-wise Results}
\noindent\textbf{Performance Stratification Consistency Across Difficulty Levels}
The consistent performance hierarchy of \textit{Human $>$ Proprietary MLLMs $>$ Random Policy $>$ Open-source MLLMs} persists uniformly across all four difficulty levels, validating our earlier observation that \textit{human} agent vastly outperforms all MLLMs while most \textit{open-source} models fail to surpass \textit{random} policy. As shown in the detailed level-wise results (tables \ref{tab:level0_results}-\ref{tab:level3_results}), this ranking remains stable as a whole despite absolute performance degradation as complexity increases.

\noindent\textbf{Limited High-Order Self-Referential Reasoning in Top Proprietary Models}
Despite relatively strong performance at levels 0 and 1 (where Claude-Sonnet-4.5 achieves 0.609 and Gemini-2.5-Pro achieves 0.585 at average scores), these leading \textit{proprietary} models exhibit dramatic performance collapse at Levels 2 and 3. This sharp decline reveals that even the most advanced MLLMs possess only rudimentary mirror-related perception but fundamentally lack high-order self-representation required for complex self-referential reasoning.

\noindent\textbf{Complete Breakdown of Task Execution at Advanced Complexity Levels}
A critical observation emerges at Levels 2 and 3: nearly all models exhibit Task Success Rates (TSRs) approaching zero across both human and robot settings. This universal failure pattern transcends model architecture, parameter scale, and training methodology, indicating that MLLMs lack robust and continuous self-representation capabilities. While some models occasionally make correct decisions at individual time steps, they fail to maintain coherent spatial reasoning throughout the entire task sequence. This fragmentation in self-referential understanding prevents models from building consistent mental maps of mirror environments, ultimately leading to complete task failure.

\begin{table*}[htbp]
\centering
\scriptsize
\setlength{\tabcolsep}{6pt}
\renewcommand{\arraystretch}{1.4}
\caption{$\mbench$ Results on Level 0. [Keys: \CLA{Highest}; \CLB{Second Highest}.]}
\label{tab:level0_results}
\vspace{-8pt}
\begin{tabular}{
l 
c c c c c 
c c c c c 
c c c c >{\columncolor{gray!20}}c
}
\toprule
\textbf{Model}
& \multicolumn{5}{c}{\textbf{Human}}
& \multicolumn{5}{c}{\textbf{Robot}}
& \multicolumn{5}{c}{\textbf{Overall}} \\
\cmidrule(lr){2-6} \cmidrule(lr){7-11} \cmidrule(lr){12-16}
& TSR & SIR & FCR & PCR & AVG
& TSR & SIR & FCR & PCR & AVG
& TSR & SIR & FCR & PCR & AVG \\
\midrule
Gemini-2.5-Pro-API
& \CLB{0.458} & \CLB{0.705} & \CLB{0.844} & \CLB{0.916} & \CLB{0.731}
& \CLA{0.611} & \CLA{0.578} & \CLA{0.546} & \CLA{0.778} & \CLA{0.628}
& \CLA{0.500} & \CLA{0.679} & \CLA{0.763} & \CLA{0.878} & \CLA{0.705} \\
Claude-Sonnet-4.5-API
& \CLA{0.542} & \CLA{0.730} & \CLA{0.891} & \CLA{0.944} & \CLA{0.777}
& 0.167 & 0.506 & 0.226 & 0.631 & 0.382
& \CLB{0.439} & \CLB{0.674} & 0.710 & \CLB{0.859} & \CLB{0.671} \\
Gemini-2.5-Flash-API
& 0.167 & 0.673 & 0.812 & 0.886 & 0.634
& \CLB{0.444} & \CLB{0.569} & \CLB{0.500} & \CLB{0.750} & \CLB{0.566}
& 0.242 & 0.654 & \CLB{0.727} & 0.849 & 0.618 \\
GLM-4.5V-106B
& 0.417 & 0.683 & 0.783 & 0.869 & 0.688
& 0.000 & 0.386 & -0.447 & 0.451 & 0.097
& 0.308 & 0.612 & 0.461 & 0.760 & 0.535 \\
GPT-4.1-API
& 0.217 & 0.650 & 0.689 & 0.792 & 0.587
& 0.111 & 0.478 & 0.078 & 0.563 & 0.307
& 0.188 & 0.609 & 0.517 & 0.728 & 0.510 \\
GPT-4o-API
& 0.208 & 0.645 & 0.671 & 0.782 & 0.577
& 0.059 & 0.416 & -0.278 & 0.525 & 0.181
& 0.169 & 0.595 & 0.423 & 0.715 & 0.476 \\
Qwen2.5-VL-72B
& 0.083 & 0.563 & 0.300 & 0.502 & 0.362
& 0.118 & 0.459 & -0.065 & 0.482 & 0.249
& 0.092 & 0.541 & 0.205 & 0.497 & 0.334 \\
Qwen-VL-Max-API
& 0.271 & 0.569 & 0.373 & 0.717 & 0.482
& 0.056 & 0.255 & -1.126 & 0.474 & -0.085
& 0.212 & 0.496 & -0.036 & 0.651 & 0.331 \\
GPT-4o-Mini-API
& 0.042 & 0.413 & -0.252 & 0.536 & 0.185
& 0.000 & 0.216 & -1.378 & 0.413 & -0.187
& 0.030 & 0.369 & -0.559 & 0.502 & 0.086 \\
Qwen2.5-VL-32B
& 0.000 & 0.286 & -0.775 & 0.458 & -0.008
& 0.000 & 0.404 & -0.341 & 0.533 & 0.149
& 0.000 & 0.312 & -0.656 & 0.478 & 0.033 \\
InternVL3.5-1B
& 0.000 & 0.152 & -0.558 & 0.083 & -0.081
& 0.000 & 0.074 & -0.302 & 0.074 & -0.038
& 0.000 & 0.135 & -0.488 & 0.081 & -0.068 \\
InternVL3.5-4B
& 0.021 & 0.198 & -1.191 & 0.092 & -0.220
& 0.000 & 0.169 & -1.647 & 0.247 & -0.308
& 0.015 & 0.192 & -1.311 & 0.132 & -0.243 \\
LLaVA-1.5-7B
& 0.000 & 0.176 & -1.376 & 0.112 & -0.272
& 0.000 & 0.142 & -2.004 & 0.228 & -0.409
& 0.000 & 0.168 & -1.547 & 0.144 & -0.309 \\
InternVL3.5-2B
& 0.000 & 0.151 & -1.458 & 0.137 & -0.292
& 0.000 & 0.149 & -1.930 & 0.207 & -0.393
& 0.000 & 0.151 & -1.587 & 0.156 & -0.320 \\
LLaVA-1.6-7B
& 0.000 & 0.098 & -1.689 & 0.144 & -0.362
& 0.000 & 0.043 & -2.511 & 0.044 & -0.606
& 0.000 & 0.085 & -1.913 & 0.117 & -0.428 \\
LLaVA-1.6-Llama3-8B
& 0.000 & 0.061 & -1.851 & 0.028 & -0.441
& 0.000 & 0.043 & -2.533 & 0.000 & -0.623
& 0.000 & 0.057 & -2.037 & 0.020 & -0.490 \\
Qwen2.5-VL-3B
& 0.000 & 0.018 & -1.984 & 0.042 & -0.481
& 0.000 & 0.000 & -2.354 & 0.000 & -0.588
& 0.000 & 0.014 & -2.091 & 0.030 & -0.512 \\
Qwen2.5-VL-7B
& 0.000 & 0.012 & -2.058 & 0.028 & -0.504
& 0.000 & 0.000 & -2.778 & 0.000 & -0.694
& 0.000 & 0.009 & -2.254 & 0.020 & -0.556 \\
\bottomrule
\end{tabular}
\vspace{-2mm}
\end{table*}

\begin{table*}[htbp]
\centering
\scriptsize
\setlength{\tabcolsep}{6pt}
\renewcommand{\arraystretch}{1.4}
\caption{$\mbench$ Results on Level 1. [Keys: \CLA{Highest}; \CLB{Second Highest}.]}
\label{tab:level1_results}
\vspace{-8pt}
\begin{tabular}{
l 
c c c c c 
c c c c c 
c c c c >{\columncolor{gray!20}}c
}
\toprule
\textbf{Model}
& \multicolumn{5}{c}{\textbf{Human}}
& \multicolumn{5}{c}{\textbf{Robot}}
& \multicolumn{5}{c}{\textbf{Overall}} \\
\cmidrule(lr){2-6} \cmidrule(lr){7-11} \cmidrule(lr){12-16}
& TSR & SIR & FCR & PCR & AVG
& TSR & SIR & FCR & PCR & AVG
& TSR & SIR & FCR & PCR & AVG \\
\midrule
GPT-4.1-API
& \CLA{0.333} & \CLB{0.663} & \CLB{0.702} & \CLB{0.767} & \CLB{0.616}
& \CLA{0.111} & \CLB{0.516} & \CLA{0.333} & \CLA{0.502} & \CLA{0.366}
& \CLA{0.273} & \CLA{0.629} & \CLA{0.601} & \CLB{0.695} & \CLA{0.550} \\
Claude-Sonnet-4.5-API
& \CLA{0.333} & \CLA{0.676} & \CLA{0.749} & \CLA{0.830} & \CLA{0.647}
& 0.000 & 0.457 & 0.161 & 0.467 & 0.271
& \CLB{0.242} & \CLB{0.623} & \CLB{0.588} & \CLA{0.731} & \CLB{0.546} \\
Gemini-2.5-Pro-API
& \CLB{0.104} & 0.646 & 0.609 & 0.722 & 0.520
& 0.056 & \CLA{0.533} & \CLB{0.185} & \CLB{0.478} & \CLB{0.313}
& 0.091 & 0.621 & 0.494 & 0.655 & 0.465 \\
GPT-4o-API
& \CLB{0.104} & 0.645 & 0.676 & 0.751 & 0.544
& 0.000 & 0.500 & 0.046 & 0.306 & 0.213
& 0.076 & 0.612 & 0.504 & 0.630 & 0.456 \\
GLM-4.5V-106B
& 0.083 & 0.580 & 0.394 & 0.591 & 0.412
& \CLB{0.059} & 0.432 & -0.067 & 0.461 & 0.221
& 0.077 & 0.548 & 0.273 & 0.557 & 0.364 \\
Gemini-2.5-Flash-API
& 0.000 & 0.578 & 0.379 & 0.507 & 0.366
& 0.056 & 0.493 & -0.030 & 0.363 & 0.220
& 0.015 & 0.559 & 0.268 & 0.468 & 0.328 \\
Qwen2.5-VL-32B
& 0.000 & 0.550 & 0.208 & 0.413 & 0.293
& 0.056 & 0.457 & -0.209 & 0.387 & 0.173
& 0.015 & 0.529 & 0.094 & 0.406 & 0.261 \\
Qwen-VL-Max-API
& 0.000 & 0.517 & 0.080 & 0.448 & 0.261
& 0.000 & 0.447 & -0.276 & 0.337 & 0.127
& 0.000 & 0.502 & -0.017 & 0.418 & 0.226 \\
InternVL3.5-4B
& 0.000 & 0.492 & 0.158 & 0.482 & 0.283
& 0.000 & 0.390 & -0.448 & 0.306 & 0.062
& 0.000 & 0.469 & -0.008 & 0.434 & 0.224 \\
Qwen2.5-VL-72B
& 0.000 & 0.540 & 0.169 & 0.397 & 0.276
& 0.000 & 0.418 & -0.444 & 0.285 & 0.065
& 0.000 & 0.513 & 0.001 & 0.367 & 0.220 \\
LLaVA-1.6-Llama3-8B
& 0.000 & 0.290 & -0.236 & 0.161 & 0.054
& 0.056 & 0.387 & -0.457 & 0.213 & 0.049
& 0.015 & 0.311 & -0.296 & 0.175 & 0.051 \\
GPT-4o-Mini-API
& 0.000 & 0.251 & -0.961 & 0.103 & -0.152
& 0.000 & 0.160 & -1.887 & 0.048 & -0.420
& 0.000 & 0.231 & -1.214 & 0.088 & -0.224 \\
Qwen2.5-VL-7B
& 0.000 & 0.212 & -1.139 & 0.129 & -0.200
& 0.000 & 0.203 & -1.671 & 0.147 & -0.330
& 0.000 & 0.210 & -1.278 & 0.133 & -0.234 \\
InternVL3.5-2B
& 0.000 & 0.174 & -1.089 & 0.042 & -0.218
& 0.000 & 0.096 & -2.156 & 0.028 & -0.508
& 0.000 & 0.156 & -1.380 & 0.038 & -0.296 \\
Qwen2.5-VL-3B
& 0.000 & 0.057 & -1.313 & 0.068 & -0.297
& 0.000 & 0.048 & -2.058 & 0.083 & -0.482
& 0.000 & 0.055 & -1.462 & 0.071 & -0.334 \\
LLaVA-1.6-7B
& 0.000 & 0.128 & -1.576 & 0.251 & -0.299
& 0.000 & 0.096 & -2.233 & 0.133 & -0.501
& 0.000 & 0.120 & -1.758 & 0.218 & -0.355 \\
InternVL3.5-1B
& 0.000 & 0.011 & -1.194 & 0.007 & -0.294
& 0.000 & 0.025 & -2.311 & 0.011 & -0.569
& 0.000 & 0.014 & -1.498 & 0.008 & -0.369 \\
LLaVA-1.5-7B
& 0.000 & 0.012 & -2.058 & 0.028 & -0.504
& 0.000 & 0.000 & -2.778 & 0.000 & -0.694
& 0.000 & 0.009 & -2.254 & 0.020 & -0.556 \\
\bottomrule
\end{tabular}
\vspace{-2mm}
\end{table*}

\begin{table*}[htbp]
\centering
\scriptsize
\setlength{\tabcolsep}{6pt}
\renewcommand{\arraystretch}{1.4}
\caption{$\mbench$ Results on Level 2. [Keys: \CLA{Highest}; \CLB{Second Highest}.]}
\label{tab:level2_results}
\vspace{-8pt}
\begin{tabular}{
l 
c c c c c 
c c c c c 
c c c c >{\columncolor{gray!20}}c
}
\toprule
\textbf{Model}
& \multicolumn{5}{c}{\textbf{Human}}
& \multicolumn{5}{c}{\textbf{Robot}}
& \multicolumn{5}{c}{\textbf{Overall}} \\
\cmidrule(lr){2-6} \cmidrule(lr){7-11} \cmidrule(lr){12-16}
& TSR & SIR & FCR & PCR & AVG
& TSR & SIR & FCR & PCR & AVG
& TSR & SIR & FCR & PCR & AVG \\
\midrule
Gemini-2.5-Pro-API
& \CLA{0.000} & \CLB{0.522} & \CLB{0.094} & \CLB{0.489} & \CLB{0.276}
& \CLA{0.000} & \CLA{0.482} & \CLA{-0.081} & \CLA{0.393} & \CLA{0.198}
& \CLA{0.000} & \CLA{0.513} & \CLA{0.046} & \CLB{0.463} & \CLA{0.256} \\
Gemini-2.5-Flash-API
& \CLA{0.000} & \CLA{0.548} & \CLA{0.259} & \CLA{0.525} & \CLA{0.333}
& \CLA{0.000} & 0.372 & -0.591 & \CLB{0.372} & 0.038
& \CLA{0.000} & \CLB{0.509} & \CLB{0.027} & \CLA{0.483} & \CLB{0.255} \\
Qwen-VL-Max-API
& \CLA{0.000} & 0.509 & 0.045 & 0.393 & 0.237
& \CLA{0.000} & 0.379 & -0.663 & 0.296 & 0.003
& \CLA{0.000} & 0.480 & -0.148 & 0.367 & 0.175 \\
GPT-4o-API
& \CLA{0.000} & 0.505 & 0.033 & 0.371 & 0.227
& \CLA{0.000} & 0.330 & -0.931 & 0.030 & -0.143
& \CLA{0.000} & 0.466 & -0.230 & 0.278 & 0.129 \\
Qwen2.5-VL-32B
& \CLA{0.000} & 0.504 & 0.013 & 0.302 & 0.205
& \CLA{0.000} & 0.319 & -1.000 & 0.102 & -0.145
& \CLA{0.000} & 0.463 & -0.263 & 0.248 & 0.112 \\
LLaVA-1.6-Llama3-8B
& \CLA{0.000} & 0.467 & -0.112 & 0.224 & 0.145
& \CLA{0.000} & 0.387 & -0.541 & 0.228 & 0.018
& \CLA{0.000} & 0.449 & -0.231 & 0.225 & 0.111 \\
Qwen2.5-VL-72B
& \CLA{0.000} & 0.464 & -0.148 & 0.292 & 0.152
& \CLA{0.000} & 0.280 & -1.202 & 0.111 & -0.203
& \CLA{0.000} & 0.423 & -0.435 & 0.243 & 0.058 \\
InternVL3.5-4B
& \CLA{0.000} & 0.380 & -0.504 & 0.120 & -0.001
& \CLA{0.000} & 0.404 & -0.381 & 0.276 & 0.075
& \CLA{0.000} & 0.385 & -0.471 & 0.162 & 0.019 \\
GPT-4o-Mini-API
& \CLA{0.000} & 0.387 & -0.481 & 0.189 & 0.024
& \CLA{0.000} & 0.309 & -1.085 & 0.056 & -0.180
& \CLA{0.000} & 0.370 & -0.646 & 0.153 & -0.031 \\
LLaVA-1.6-7B
& \CLA{0.000} & 0.267 & -0.912 & 0.064 & -0.145
& \CLA{0.000} & \CLB{0.465} & \CLB{-0.200} & 0.172 & \CLB{0.109}
& \CLA{0.000} & 0.312 & -0.715 & 0.094 & -0.077 \\
GPT-4.1-API
& \CLA{0.000} & 0.373 & -0.541 & 0.127 & -0.010
& \CLA{0.000} & 0.227 & -1.511 & 0.074 & -0.303
& \CLA{0.000} & 0.340 & -0.805 & 0.113 & -0.088 \\
Qwen2.5-VL-7B
& \CLA{0.000} & 0.301 & -0.793 & 0.200 & -0.073
& \CLA{0.000} & 0.245 & -1.444 & 0.119 & -0.270
& \CLA{0.000} & 0.288 & -0.970 & 0.178 & -0.126 \\
Claude-Sonnet-4.5-API
& \CLA{0.000} & 0.297 & -0.839 & 0.116 & -0.107
& \CLA{0.000} & 0.188 & -1.687 & 0.046 & -0.363
& \CLA{0.000} & 0.273 & -1.071 & 0.097 & -0.175 \\
GLM-4.5V-106B
& \CLA{0.000} & 0.270 & -0.918 & 0.063 & -0.146
& \CLA{0.000} & 0.156 & -1.929 & 0.021 & -0.438
& \CLA{0.000} & 0.247 & -1.171 & 0.052 & -0.218 \\
InternVL3.5-2B
& \CLA{0.000} & 0.226 & -0.976 & 0.119 & -0.158
& \CLA{0.000} & 0.124 & -2.007 & 0.046 & -0.459
& \CLA{0.000} & 0.203 & -1.257 & 0.099 & -0.239 \\
Qwen2.5-VL-3B
& \CLA{0.000} & 0.105 & -1.334 & 0.067 & -0.291
& \CLA{0.000} & 0.096 & -1.725 & 0.083 & -0.386
& \CLA{0.000} & 0.103 & -1.412 & 0.070 & -0.310 \\
InternVL3.5-1B
& \CLA{0.000} & 0.027 & -1.271 & 0.014 & -0.307
& \CLA{0.000} & 0.046 & -2.224 & 0.009 & -0.542
& \CLA{0.000} & 0.032 & -1.531 & 0.013 & -0.372 \\
LLaVA-1.5-7B
& \CLA{0.000} & 0.012 & -1.843 & 0.028 & -0.451
& \CLA{0.000} & 0.000 & -2.778 & 0.000 & -0.694
& \CLA{0.000} & 0.009 & -2.098 & 0.020 & -0.517 \\
\bottomrule
\end{tabular}
\vspace{-2mm}
\end{table*}

\begin{table*}[htbp]
\centering
\scriptsize
\setlength{\tabcolsep}{6pt}
\renewcommand{\arraystretch}{1.4}
\caption{$\mbench$ Results on Level 3. [Keys: \CLA{Highest}; \CLB{Second Highest}.]}
\label{tab:level3_results}
\vspace{-8pt}
\begin{tabular}{
l 
c c c c c 
c c c c c 
c c c c >{\columncolor{gray!20}}c
}
\toprule
\textbf{Model}
& \multicolumn{5}{c}{\textbf{Human}}
& \multicolumn{5}{c}{\textbf{Robot}}
& \multicolumn{5}{c}{\textbf{Overall}} \\
\cmidrule(lr){2-6} \cmidrule(lr){7-11} \cmidrule(lr){12-16}
& TSR & SIR & FCR & PCR & AVG
& TSR & SIR & FCR & PCR & AVG
& TSR & SIR & FCR & PCR & AVG \\
\midrule
Gemini-2.5-Pro-API
& \CLA{0.000} & \CLB{0.571} & \CLB{0.302} & \CLB{0.540} & \CLB{0.353}
& \CLA{0.111} & \CLA{0.519} & \CLA{0.122} & \CLA{0.485} & \CLA{0.309}
& \CLA{0.030} & \CLA{0.560} & \CLA{0.253} & \CLA{0.525} & \CLA{0.342} \\
Gemini-2.5-Flash-API
& \CLA{0.000} & 0.537 & 0.164 & 0.509 & 0.303
& \CLB{0.000} & \CLB{0.390} & \CLB{-0.461} & \CLB{0.261} & \CLB{0.048}
& \CLB{0.000} & 0.504 & \CLB{-0.006} & \CLB{0.441} & \CLB{0.235} \\
Qwen-VL-Max-API
& \CLA{0.000} & \CLA{0.577} & \CLA{0.336} & \CLA{0.546} & \CLA{0.365}
& \CLB{0.000} & 0.309 & -1.052 & 0.054 & -0.172
& \CLB{0.000} & \CLB{0.517} & -0.042 & 0.412 & 0.222 \\
GPT-4.1-API
& \CLA{0.000} & 0.497 & -0.014 & 0.439 & 0.231
& \CLB{0.000} & 0.266 & -1.269 & 0.085 & -0.229
& \CLB{0.000} & 0.445 & -0.356 & 0.343 & 0.108 \\
InternVL3.5-4B
& \CLA{0.000} & 0.433 & -0.206 & 0.342 & 0.142
& \CLB{0.000} & 0.352 & -0.814 & 0.088 & -0.093
& \CLB{0.000} & 0.416 & -0.365 & 0.276 & 0.082 \\
Qwen2.5-VL-72B
& \CLA{0.000} & 0.421 & -0.333 & 0.327 & 0.104
& \CLB{0.000} & 0.259 & -1.333 & 0.028 & -0.262
& \CLB{0.000} & 0.385 & -0.606 & 0.245 & 0.006 \\
Qwen2.5-VL-32B
& \CLA{0.000} & 0.430 & -0.292 & 0.169 & 0.077
& \CLB{0.000} & 0.177 & -1.793 & 0.009 & -0.402
& \CLB{0.000} & 0.374 & -0.702 & 0.126 & -0.051 \\
GPT-4o-API
& \CLA{0.000} & 0.409 & -0.371 & 0.112 & 0.038
& \CLB{0.000} & 0.216 & -1.472 & 0.028 & -0.307
& \CLB{0.000} & 0.364 & -0.686 & 0.088 & -0.058 \\
Claude-Sonnet-4.5-API
& \CLA{0.000} & 0.307 & -0.800 & 0.152 & -0.085
& \CLB{0.000} & 0.206 & -1.596 & 0.020 & -0.343
& \CLB{0.000} & 0.284 & -1.017 & 0.116 & -0.154 \\
LLaVA-1.6-7B
& \CLA{0.000} & 0.239 & -1.101 & 0.229 & -0.158
& \CLB{0.000} & 0.103 & -2.182 & 0.111 & -0.492
& \CLB{0.000} & 0.209 & -1.395 & 0.197 & -0.248 \\
GLM-4.5V-106B
& \CLA{0.000} & 0.229 & -1.068 & 0.120 & -0.180
& \CLB{0.000} & 0.060 & -2.387 & 0.019 & -0.577
& \CLB{0.000} & 0.191 & -1.434 & 0.092 & -0.288 \\
Qwen2.5-VL-7B
& \CLA{0.000} & 0.209 & -1.141 & 0.086 & -0.211
& \CLB{0.000} & 0.057 & -2.357 & 0.028 & -0.568
& \CLB{0.000} & 0.175 & -1.473 & 0.070 & -0.307 \\
Qwen2.5-VL-3B
& \CLA{0.000} & 0.091 & -1.379 & 0.031 & -0.314
& \CLB{0.000} & 0.005 & -1.972 & 0.014 & -0.488
& \CLB{0.000} & 0.078 & -1.497 & 0.028 & -0.348 \\
GPT-4o-Mini-API
& \CLA{0.000} & 0.137 & -1.381 & 0.026 & -0.304
& \CLB{0.000} & 0.021 & -2.507 & 0.000 & -0.622
& \CLB{0.000} & 0.110 & -1.697 & 0.018 & -0.392 \\
InternVL3.5-1B
& \CLA{0.000} & 0.016 & -1.331 & 0.000 & -0.329
& \CLB{0.000} & 0.011 & -2.341 & 0.000 & -0.583
& \CLB{0.000} & 0.015 & -1.606 & 0.000 & -0.398 \\
InternVL3.5-2B
& \CLA{0.000} & 0.039 & -1.431 & 0.004 & -0.347
& \CLB{0.000} & 0.032 & -2.461 & 0.009 & -0.605
& \CLB{0.000} & 0.037 & -1.712 & 0.005 & -0.417 \\
LLaVA-1.6-Llama3-8B
& \CLA{0.000} & 0.016 & -1.954 & 0.021 & -0.479
& \CLB{0.000} & 0.046 & -2.437 & 0.000 & -0.598
& \CLB{0.000} & 0.023 & -2.086 & 0.015 & -0.512 \\
LLaVA-1.5-7B
& \CLA{0.000} & 0.012 & -2.058 & 0.028 & -0.504
& \CLB{0.000} & 0.000 & -2.778 & 0.000 & -0.694
& \CLB{0.000} & 0.009 & -2.254 & 0.020 & -0.556 \\
\bottomrule
\end{tabular}
\vspace{-2mm}
\end{table*}

\end{document}